\documentclass{article}

\usepackage{arxiv}

\usepackage[utf8]{inputenc} 
\usepackage[T1]{fontenc}    
\usepackage{hyperref}       
\usepackage{url}            
\usepackage{booktabs}       
\usepackage{amsfonts}       
\usepackage{nicefrac}       
\usepackage{microtype}      
\usepackage{lipsum}		
\usepackage{graphicx}
\usepackage{doi}

\usepackage{xcolor} 
 \usepackage{array,multirow,graphicx} 
 \usepackage[caption=false]{subfig} 
 \usepackage{amsmath}  
 \usepackage{booktabs}  

\usepackage{algorithm}
\usepackage{algorithmicx}
 \usepackage{algpseudocode}

\title{Deep spatial context: when attention-based models meet spatial regression}

\author{ \href{https://orcid.org/0000-0001-6767-1018}{\includegraphics[scale=0.06]{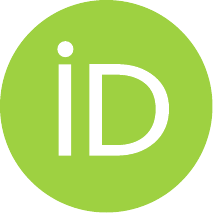}\hspace{1mm}Paulina~Tomaszewska}\\
	Faculty of Mathematics and Information Science\\
	Warsaw University of Technology\\
	Poland \\
	\And
    	\href{https://orcid.org/0000-0001-9468-4674}{\includegraphics[scale=0.06]{orcid.pdf}\hspace{1mm}Elżbieta Sienkiewicz} \\
	Faculty of Mathematics and Information Science\\
	Warsaw University of Technology\\
	Poland \\
        \And
        {Mai P. Hoang} \\
	Department of Pathology \\
    Massachusetts General Hospital and Harvard Medical School\\
    Boston, USA \\
    \And
    \href{https://orcid.org/0000-0001-8423-1823}{\includegraphics[scale=0.06]{orcid.pdf}\hspace{1mm}Przemysław Biecek} \\
	Faculty of Mathematics and Information Science\\
	Warsaw University of Technology\\
	Poland \\
}

\date{}

\renewcommand{\shorttitle}{Deep spatial context: when attention-based models meet spatial regression}

\hypersetup{
pdftitle={Deep spatial context: when attention-based models meet spatial regression},
}

\begin{document}
\maketitle

\begin{abstract}
We propose \emph{`Deep spatial context'} (\emph{DSCon}) method, which serves for investigation of~the~attention-based vision models using the concept of spatial context. It was inspired by histopathologists, however, the method can be applied to various domains. 
The \emph{DSCon} 
allows for a quantitative measure of the spatial context's role using three \underline{S}patial \underline{C}ontext \underline{M}easures: $SCM_{features}$, $SCM_{targets}$, $SCM_{residuals}$ to distinguish whether the spatial context is observable within the features of neighboring regions, their target values (attention scores) or residuals, respectively. It is achieved by integrating spatial regression into the pipeline.
The \emph{DSCon} helps to verify research questions. The experiments reveal that spatial relationships are much bigger in the case of the classification of tumor lesions than normal tissues. Moreover, it turns out that the larger the size of the neighborhood taken into account within spatial regression, the less valuable contextual information is. Furthermore, it is observed that the spatial context measure is the largest when considered within the feature space as opposed to the~targets and residuals.
\end{abstract}

\keywords{spatial context \and explainable AI \and spatial statistics \and digital pathology}

\section{Introduction}
The behavior and reasoning of Deep Learning (DL) models are often hard to understand, especially for people without experience in artificial intelligence (AI). This is the source of a low level of trust in such models, especially in critical domains, e.g. healthcare. Physicians would like to know if DL models take similar issues into account when making a diagnosis as they do. The need to understand the models was one of the motivations behind the eXplainable AI (XAI) raise~\cite{xai_survey}. However, as pointed out in~\cite{manifesto}, most existing XAI methods do not operate on the same level of abstraction as people interested in the understanding of systems. For instance, gradient-based methods~\cite{gradcam,ig} can highlight particular pixels within images, e.g. constituting face, but this is not clear whether it is the shape, texture, color, etc. that matters. That is why we need to operate on semantic human-understandable concepts when explaining models. It is best if the concepts are specified by experts from the domain which the system is designed for. The issue was pointed out in the manifesto of XAI 2.0 (human-centered explanations)~\cite{manifesto}.

Following this notion, we collaborate with a histopathologist to learn what concepts are used by them in their daily work when diagnosing lesions. It turns out that they analyze tissues from both global (tissue-level) and local (cell-level) perspectives. This concept can be formalized as a spatial context, which is a general term and can be observed in various domains e.g. within satellite images.

\begin{figure}[h!]
\centering
\includegraphics[width=0.72\textwidth]
{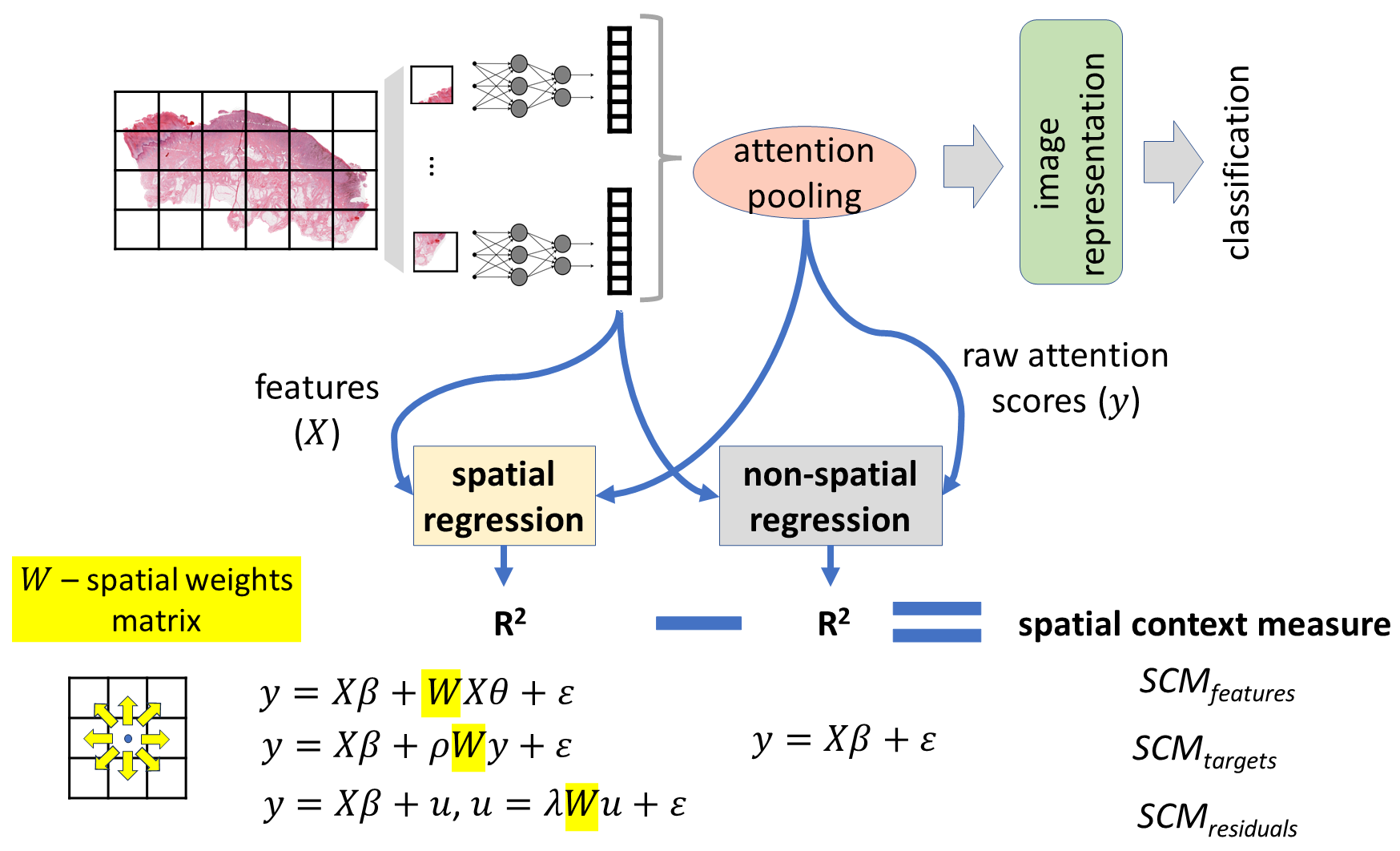}
\caption{Simplified scheme of the proposed \textit{`Deep spatial context'} method where information extracted from the~attention-based model is used within spatial regression models to assess and measure the~importance of spatial context.}
\label{fig:scheme}
\end{figure}

\paragraph{Spatial context}
The topic of context within the input data is mainly considered in time series and natural language processing, where it is understood as the number of data pieces surrounding the analyzed data sample. The fact that the size of the context matters is outlined in~\cite{LongLLama}. The concept of context in the case of images is not widely covered. It can be perceived as additional tabular information attached to the images. However, in this work, we focus on spatial context within images which we define as the fact that the location of particular elements of the image and their relative positions are significant in a decision-making process. For instance, the fact of how much the areas with tumor are spread is an important factor when assessing the disease stage. Therefore, it translates to the risk stratification and treatment~\cite{widespread}. The other illustrative example to give the intuition behind the idea of spatial context is the scenario where there are two people in an image and one has a knife. Then, there is a big difference in whether the person with a knife is heading another person or going in the opposite direction. Therefore, the relative position of the two entities matters. \\

The goal of this work is to analyze attention-based DL models from a spatial context point of view. For the illustrative domain of the application of the proposed method, we check whether the behavior of existing models in digital pathology is consistent with the daily practice of histopathologists.

The common approach to the classification of histopathological images in the form of Whole Slide Images is attention-based models. One of such solutions is CLAM~\cite{clam}. In this case, the image of huge resolution is first split into patches that are the input to the feature extractor. The features are then aggregated in the pooling module, where the assigned attention scores serve as coefficients in a weighted sum to form an overall representation of the entire original image. Such~a~representation is an input to the final classifier. 
In such attention-based models, the information about the~patches' location within the whole image is not explicitly preserved, as no positional encoding is applied. This can be troublesome in the case of histopathological images that have non-standardized sizes. 

To verify if the information on spatial context is preserved within the attention-based models, we propose to investigate the relationship between the patch features and the assigned attention scores. The model itself is not explicitly asked to learn spatial relations. However, the~fact that it achieves decent classification performance suggests that some contextual information may be captured by the model. \\

We propose a \textit{`Deep spatial context'} method (\emph{DSCon}) (Figure~\ref{fig:scheme}) which is in line with the notion of XAI 2.0~\cite{manifesto} as it investigates the DL model from the point of view of the semantic concept of spatial context. The research was inspired by histopathologists, although the method is general and can be applied to other domains. In the solution, we use spatial regression models to investigate the DL model's internals. 
\newpage
Our main contribution is as follows:
\begin{itemize}
    \item we propose a novel \emph{`Deep spatial context'} method that allows for the investigation of DL models from the perspective of spatial context concept. The method is of integrative nature by incorporating spatial regression into the explanability area. 
    \item we verify whether the spatial context is preserved in attention-based models and we provide a method to assess it quantitatively using three spatial context measures ($SCM_{features}$, $SCM_{targets}$, $SCM_{residuals}$) that enable to distinguish where the spatial relationships are maintained (feature, label, residual space)
    \item we investigate several hypotheses using the \emph{DSCon} in an illustrative digital pathology domain i.e. we check whether the spatial aspect is more important in the case of tumor lesions than the normal tissues
\end{itemize}

The code of the experiments is provided at \texttt{https://github.com/ptomaszewska/DSCon.git}.

\section{Related works}
Many works in the XAI field focus on highlighting the parts of the input image that lead to the largest neuron activations and therefore influence the final model decisions the most. In the early works, the contribution of single input pixels (saliency maps~\cite{saliency}) and superpixel (groups of pixels)~\cite{lime} were mainly developed. The recent works focus more on high-level, complex semantic concepts~\cite{tcav,crp,segment}. There are also neural networks that are explainable-by-design providing trainable prototypes~\cite{protopnet,protomil}. 
In all these solutions, the process is that the XAI methods return some explanations that require humans to name them as human-understandable, semantic concepts i.e. beak, eye. 

In our work, we reverse the process. We consult with a domain expert about what concepts are meaningful in a particular field and design a method to verify whether the concept is kept within the neural network.

In the papers from the field of digital pathology, the analysis of the model behavior is often limited to information about the global metrics like accuracy and AUC accompanied by few heatmaps with patches' importance in the overall tissue classification. There are only a few attempts to explain the models in more detail (based on prototypes~\cite{protomil} and an idea similar to global centroids~\cite{end2end}).
To the best of our knowledge, there are no works that
analyse the models from the spatial context perspective. 

The attention mechanism is widely used in different forms in DL systems, however, it was pointed out that the attention scores themselves are not valid model explanations~\cite{attention_not_enough}.
There are recent works where the attention mechanism is used in another manner to explain models. The idea is to insert the external attention mechanism into the already trained convolutional model to train it to derive class activation maps being a form of model explanation~\cite{attention,tame}.\\
However, there have been only a few attempts to investigate the way the attention mechanism works from a more theoretic point of view. The self-attention can be considered as a marginal likelihood over possible network structures~\cite{probabilistic}, via a dynamical system perspective~\cite{dynamic} or through the lens of exchangeability~\cite{exchange}.
To the best of our knowledge, we are the first to apply spatial statistics for the analysis of attention-based models and Deep Learning models, in general. However, there are already XAI methods that inherit from different mathematical subdisciplines - e.g. SHAP values~\cite{shap} are inspired by game theory.

\section{Method}
\label{sec:method}

\subsection{Preliminaries - spatial regression}
Ordinary least squares regression (OLS) ($y=X\beta + \epsilon$) is applicable when observations and residuals are independent of each other (there are no spatial relations). However, patches constitute WSI and therefore there should be a continuity of features among them. Moreover, when analyzing the particular patch, the information from the neighboring patches could be valuable to take into account (which is consistent with the histopathologists' expertise).

The two common spatial regression classes of models are (1) spatial lag model ($y=\rho(Wy)+W\beta+\epsilon$) i.e in a form of spatial two-stage least squares 
model~\cite{regression} (denoted as \textit{lag} method for brevity), (2) spatial error model ($y=X\beta + u$ where $u=\lambda(Wu)+\epsilon$) i.e in a form of generalized method of moments for a spatial error model~\cite{error1,error2} (denoted as \textit{error} method). 
Moreover, the OLS model can be applied with manually lagged features based on the neighboring patches - SLX ($y=X\beta+WX\theta+\epsilon$) (referred as \textit{Wx}). Lagging means that we use a transformation to incorporate the information from the neighboring areas (more in the Appendix). This model is actually `pseudo-spatial'; however, we will not use this distinction in the following sections. \\
In the spatial models, the information about the analyzed neighborhood is provided in the spatial weights matrix $W$. It can be defined using Rook's or Queen's criterion. In the former, the neighbors are the regions that share a border, whereas in the latter, having at least one common point is enough. Alternatively, the distance-band approach can be taken i.e. by using \emph{k}-nearest neighbors. Note that the three analyzed spatial methods (\textit{Wx}, \textit{lag}, \textit{error}) vary in terms of where within the regression formula the~spatial component ($W$) is applied.

\subsection{\textit{Deep Spatial Context}}
First, let us formalize the attention-based model. We are given an image of huge resolution (\textit{input image}) which is then split into \emph{N} patches 
after removing the background. The~patches are an independent input to the feature extractor that returns the features $h_i$ (for $i=1,..,N$). The features are passed to the attention pooling mechanism, which returns raw attention scores ($a$) that are later transformed into the probability space using the softmax function ($a^{softmax}$). The overall representation of the \textit{input image} defined as ${z=\sum_{i=1}^{N} a_i^{softmax}*h_i}$ is passed to feed-forward layers for the final classification.

The key mechanism of the attention-based model assigns attention scores to patch features. To understand this relationship from the spatial context point of view, we propose the~\emph{`Deep spatial context'} (\emph{DSCon}) method incorporating spatial regression. 
The goal of the proposed method is to capture the difference in regression performance when the~spatial component is included within the regression model versus the case without it; therefore, to assess the importance of spatial context. 

The scheme of the \emph{DSCon} method is provided in Algorithm~\ref{alg:algorithm}. The first step, when having an already trained attention-based model (i.e. CLAM), is to perform inference and save the extracted patch features $h_i$ and raw attention scores $a_i$. 
The raw attention scores do not have the interpretation of probabilities unlike $a^{softmax}$ but their distribution is closer to Gaussian, which is desirable in regression models. 
Before, the regression model is going to be performed on the features ($H$) as input and raw attention scores as target, it is necessary to filter out the elements of features causing collinearity that is detrimental to the regression models. The filtering is done based on the Pearson correlation (given a threshold).
Next, the OLS regression (without any spatial component) is run for each \textit{input image} separately to predict the raw attention scores ($a$) based on a set of features after filtering. The~resulting performance metric of OLS is saved (in the analyzed example due to the different WSIs' sizes, it is $R^2$). This value will serve as a reference point for further analysis.

The next step is to check whether the spatial regression is applicable. The common indicator is that the residuals of OLS form spatial clusters. It can be formally verified by performing Global Moran’s I test~\cite{global_moran} which checks if there is a global spatial autocorrelation in the input data (see Appendix). At this stage, the spatial weights matrix ($W$) has to be defined i.e. using $k$-nearest neighbors. Later, it is checked whether the $p$-value returned in the test is smaller than the defined significance level. If so, the application of spatial regression is justifiable. 
Using the same set of features as in OLS, spatial regression models (\emph{Wx}, \emph{lag}, \emph{error}) are run and the returned performance metrics are saved. 

The final step is to compute the difference between the performance metrics obtained using each spatial model (\emph{Wx}, \emph{lag}, \emph{error}) and OLS; as a result, we get the \underline{S}patial \underline{C}ontext \underline{M}easures: $SCM_{features}$, $SCM_{targets}$ and $SCM_{residuals}$, respectively.
If the spatial context metric is positive, it means that the consideration of the spatial context in a particular part of the regression model is beneficial meaning that the spatial relationships are maintained in the trained DL model. However, if the metric is negative, it means that the incorporation of spatial information introduces too much noise to the regression model as the `spatial signal' is not present in the model. The value of the spatial context measure gives quantitative information about how important the spatial context~is.

\begin{algorithm}[tb]
\caption{\textit{Deep spatial context (DSCon)}}
\label{alg:algorithm}
\textbf{Input}: patch features ($h_i$), raw attention scores assigned to patches ($a_i$)\\
\textbf{Parameter}: correlation threshold ($\gamma$), neighborhood size~($k$)\\
\textbf{Output}: spatial context measures ($SCM_{features}$, $SCM_{targets}$, $SCM_{residuals}$)

\begin{algorithmic}
\For {$img$ in $dataset$}
\State $H \gets \{h_1; ...; h_n\}$
\State $a \gets \{a_1, ..., a_n\}$

\State {$c \gets \{correlation (H_{:,i}, H_{:,j})\}$ }
\If {$abs(c_{ij})>\gamma$}
\State {drop $H_{:,i}$}  \Comment{Reduce multicollinearity by dropping correlated features}
\EndIf



\State $R^2_{OLS}, residuals_{OLS} \gets OLS(H, a)$ \Comment{Run OLS regression}
\State $W \gets spatial\_weights(type=kNN, k)$ \Comment{Define spatial weights matrix}
\State $p_{Moran} \gets Moran\_test(W, residuals_{OLS})$   \Comment{Check if there is a spatial autocorrelation}
\If {$p_{Moran}>significance\_level$}
\State skip
\EndIf
\State $R^2_{Wx} \gets Wx(H, a, W)$   \Comment{Run spatial regression models}
\State $R^2_{lag} \gets lag(H, a, W)$
\State $R^2_{error}, \gets error(H, a, W)$
\State
\State $SCM_{features} \gets R^2_{Wx} - R^2_{OLS}$ \Comment{Calculate spatial context measures}
\State $SCM_{targets} \gets R^2_{lag} - R^2_{OLS}$
\State $SCM_{residuals} \gets R^2_{error} - R^2_{OLS}$

\EndFor
\end{algorithmic}
\end{algorithm}

\section{Experiments}
\subsection{Deep Learning models}
In all illustrative experiments, we used the Camelyon16 Breast Cancer dataset~\cite{camelyon} containing 399 WSIs in total.
Inspired by the Rashomon's quartet ~\cite{rashomon}, where several models of similar performance proved to reason about the data in different ways, we decided to test the proposed \textit{`Deep spatial context'} method on a range of different attention-based models achieving similar classification performance to verify whether, in this case, the~differences in capturing the spatial context are significant. 

In the experiments, we use CLAM implementation of the~attention-based model~\cite{clam}. We replaced the baseline \emph{Resnet50} feature extractor with different transformer-based backbones to verify how the extracted features vary and later impact the learned attention scores which are the core input to the \emph{DSCon}. We applied the \emph{Swin} model~\cite{swin} also in the second version~\cite{swinv2} which are transformer models that resemble the convolutional models due to a hierarchical structure. Furthermore, we used \emph{Vision Transformer (ViT)}~\cite{vit}. Models from \emph{Swin} and \emph{ViT} families differ in terms of the scope of attention - in the first case, it is local (window-based), whereas, in the second it is global (self-attention).

\subsection{Training}
We used the convolutional and transformer-based backbones already pretrained in a supervised manner on Imagenet. Following the original CLAM setup, we did not fine-tune them on histopathological data. 

The following models served as feature extractors: \emph{Swin (base)} (\emph{patch\_size=4, window\_size=7}), \emph{Swin~v2} (\emph{tiny, small, base} with \emph{patch\_size=4} and \emph{window\_size=8} or \emph{16}), \emph{ViT (base)}. 
The extracted patch features served as an input to the attention pooling and consecutive classification layers which were trained from scratch.

\begin{figure}[h]
\centering
\includegraphics[scale=0.4]{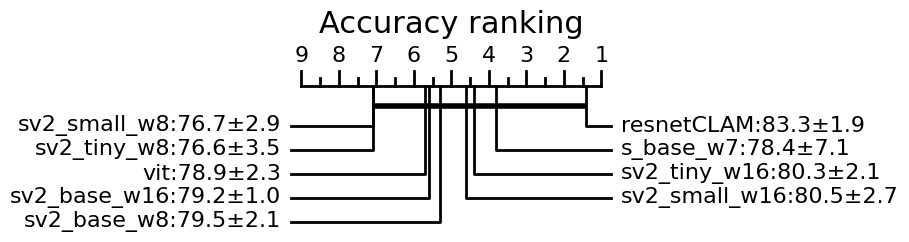}
\caption{Critical distance plot\protect\footnotemark of models' accuracy of classification of normal tissue vs. tumor lesion. For brevity: \textit{s} denotes \textit{Swin}, \textit{sv2} - \textit{Swin v2} and \textit{w} - \textit{window}. In the labels, the mean accuracy (percentage) and the standard deviation over 5 trained models during the cross-validation are provided. The fact that the horizontal line below the ranking axis connects all the models means that there is no significant difference in their performance based on the Wilcoxon-Holm method used to detect pairwise significance.
}
\label{fig:model_perf}
\end{figure}

The initial parameters of the grid search were inspired by the ones used in~\cite{vpt}. Later, they were modified based on the intermediate results. After an extensive search over many hyperparameters, we chose those that for every model variant (with different backbones) resulted in a decent performance similar to the baseline with \emph{Resnet50} backbone (denoted as \textit{resnetCLAM}) (details in the Appendix). In \textit{resnetCLAM}, the parameters claimed in the original paper were applied. To conclude, the models used in the analysis varied in terms of architecture only by the choice of the backbone. For brevity, the whole trained attention-based models (CLAM) are named after the feature extractor used. The classification performance of the models is summarized in Figure~\ref{fig:model_perf}. The differences in performance among the models are not statistically significant.
\footnotetext{The code used for the generation of the plot is from~\cite{critical}.}

\subsection{Application of \textit{`Deep spatial context'} method} 
Depending on the type of backbone used, the length of the extracted features was 768 or 1024. Subsequently, the size of the features was reduced due to filtering based on the Pearson correlation coefficients with the threshold ($\gamma=0.6$). 
Later, Moran's~I test was performed with $\alpha=0.05$. It turned out that in 99.4\% of all cases, there was a statistically significant spatial autocorrelation which justified the use of spatial regression. Here, one case is a test image analyzed using the data (features and attention scores) extracted from a particular DL model using a regression model with a spatial weights matrix defined using $k$-nearest neighbors. In short, it means that one test image was processed using the particular `model-$k$ combination'.
In the experiments, a different range of neighborhood is considered within the spatial weights matrix defined using \textit{k}-nearest neighbours (with \emph{k=8, 24, 48, 99}). In the case of WSIs, the variant with \emph{k = 8} leads to the same results as the Queen's contiguity matrix since the grid of patches is regular.

\section{Results}
The regression models were run on features obtained using different backbones. Five different raw attention scores corresponded to the features extracted from each patch using a particular backbone
as a result of a 5-fold cross-validation run on whole CLAM models. 
Therefore, in total, a particular regression model was fitted 5 times on a set of features and attention scores from a particular image. The performance results ($R^2$) were aggregated using concatenation to have more data points for the analysis (computing the mean of the results from the folds led to similar conclusions). The~results below are obtained on the test set data.\\

The \textit{`Deep spatial context'} method helps to verify different research hypotheses regarding the importance of spatial context.

\subsection{Spatial context in tumor lesions vs. normal tissues}

\begin{figure}[h!]
\centering

\subfloat[$SCM_{features}$]{%
  \includegraphics[,width=0.52\columnwidth]{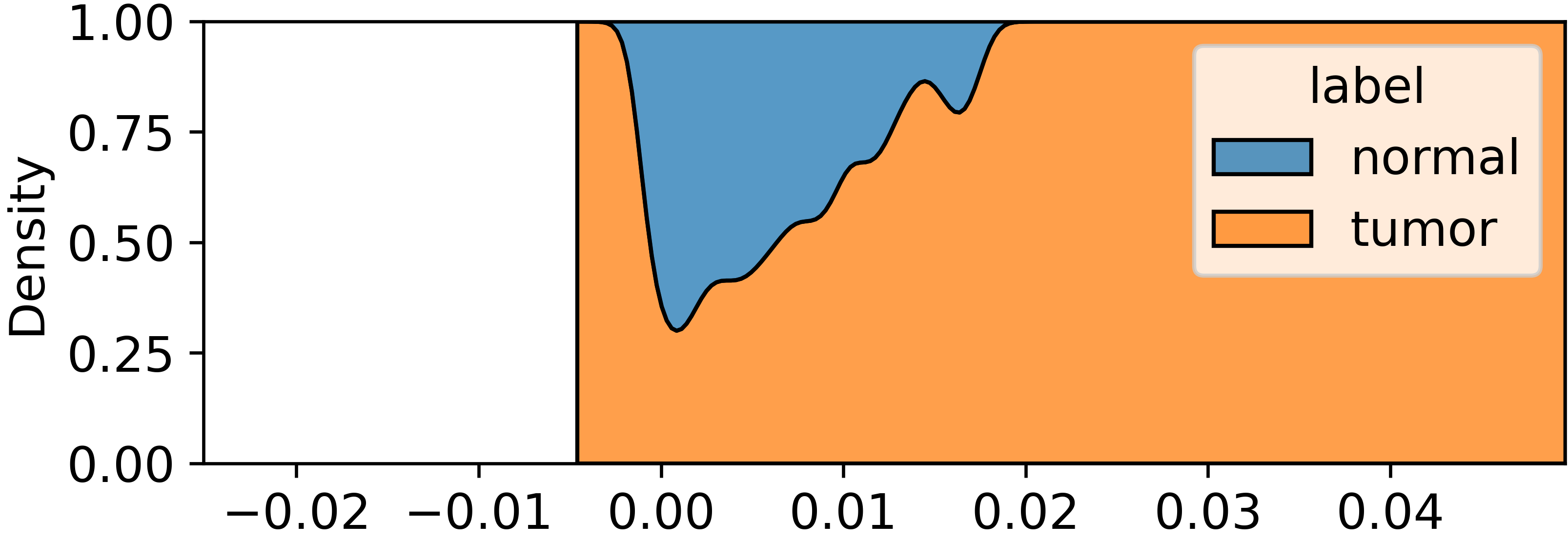 }%
}
\\[-0.5ex] 
\subfloat[$SCM_{targets}$]{%
  \includegraphics[,width=0.52\columnwidth]{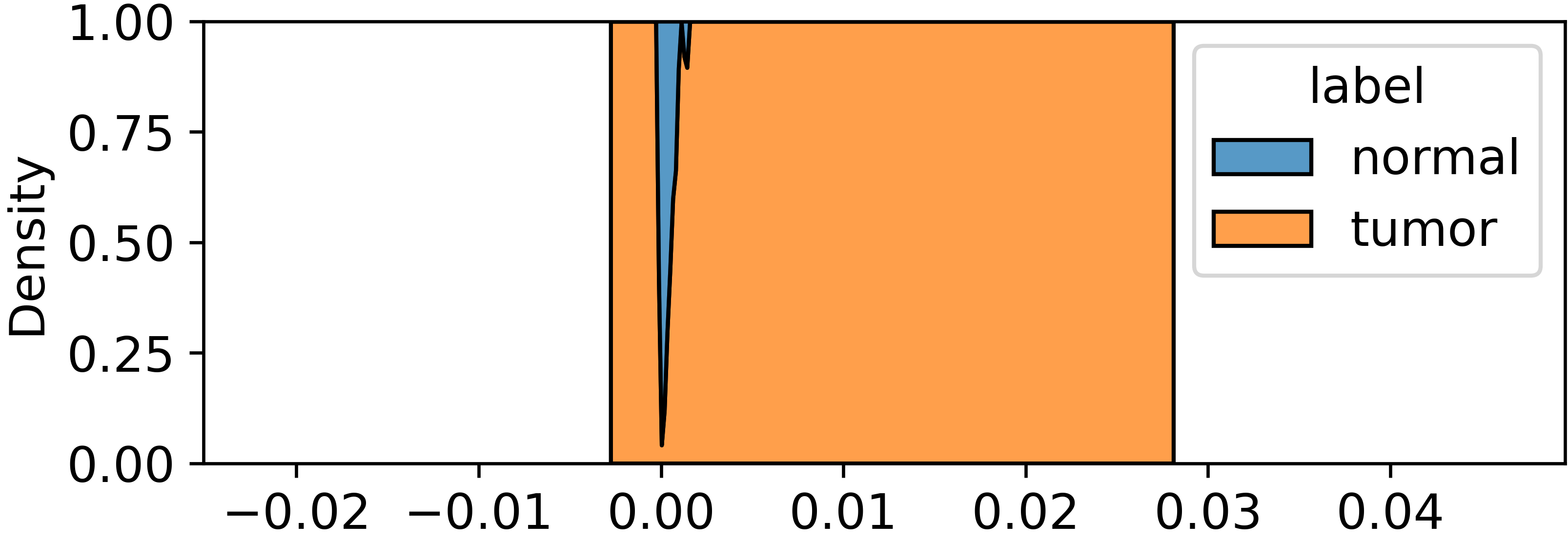}%
}
\\[-0.5ex] 
\subfloat[$SCM_{residuals}$]{%
  \includegraphics[,width=0.52\columnwidth]{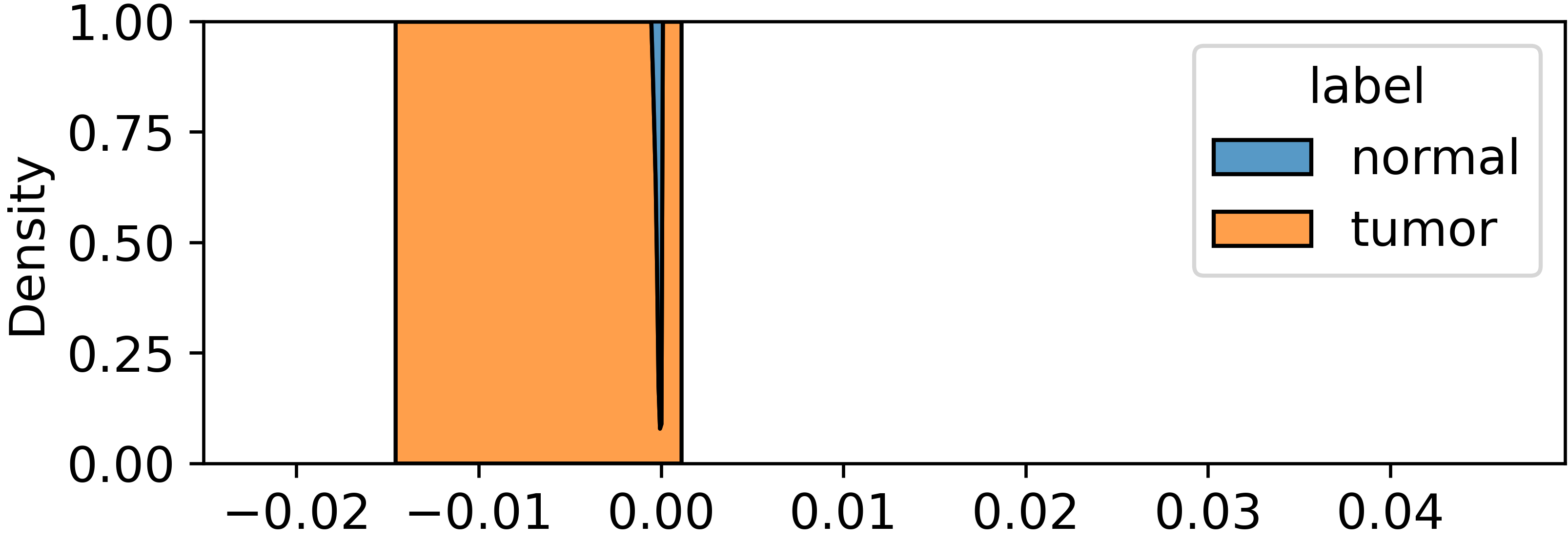}%
}
\caption{Stacked kernel distribution estimation plots where the proportion of values from normal tissues and tumor lesions for a given spatial context measure is shown. The results are for the \textit{sv2\_base\_w8} model with \textit{k=24} nearest neighbors within the spatial weights matrix. The appearance of the plots is representative of other model-$k$ combinations. The differences between the models are summarized in Table~\ref{tab:ranking}. 
}
\label{fig:density}
\end{figure}

The analysis of results from all model-$k$ combinations (shown for one variant in Figure~\ref{fig:density}) indicates that the boost of the $R^2$ metric due to spatial regression application reaches greater values in the case of tumor lesions than in normal tissues (for $SCM_{features}$ and $SCM_{targets}$). However, the range of values attributed to the normal lesions is much wider in the case of $SCM_{features}$ than $SCM_{targets}$ where the values are almost zero for normal tissues meaning that the spatial context within targets is in this case negligible. 
Moreover, $SCM_{residuals}$ almost always has negative values, which is rare in the case of $SCM_{features}$ and $SCM_{targets}$. 
The~visible differences in the boost ranges of $R^2$ in different regression models are statistically significant regardless of $k$ value within the spatial weights matrix which was verified using a~t-test with $\alpha=0.05$.

\subsubsection{Incorporating spatial aspects in different components within regression} Note that in the case of \emph{Wx} regression, the spatial weights matrix is applied only to features. 
Probably, incorporating the spatial component to targets (\textit{lag} regression) introduces some noise that degrades the mean performance gain, however, increases the separability between the normal and tumor data (see Figure~\ref{fig:density}). Even more distortion introduces the imposition of spatial context on residuals that leads to $SCM_{residuals}$ being almost always negative. As a consequence, in the following analyses, we focus on the $R^2$ boost achieved using the $Wx$ and $lag$ regression models.

\subsection{Impact of size of neighborhood (\emph{k}) 
on the spatial context measures}

\begin{figure}[h!]
\centering
\includegraphics[,width=0.5\columnwidth]{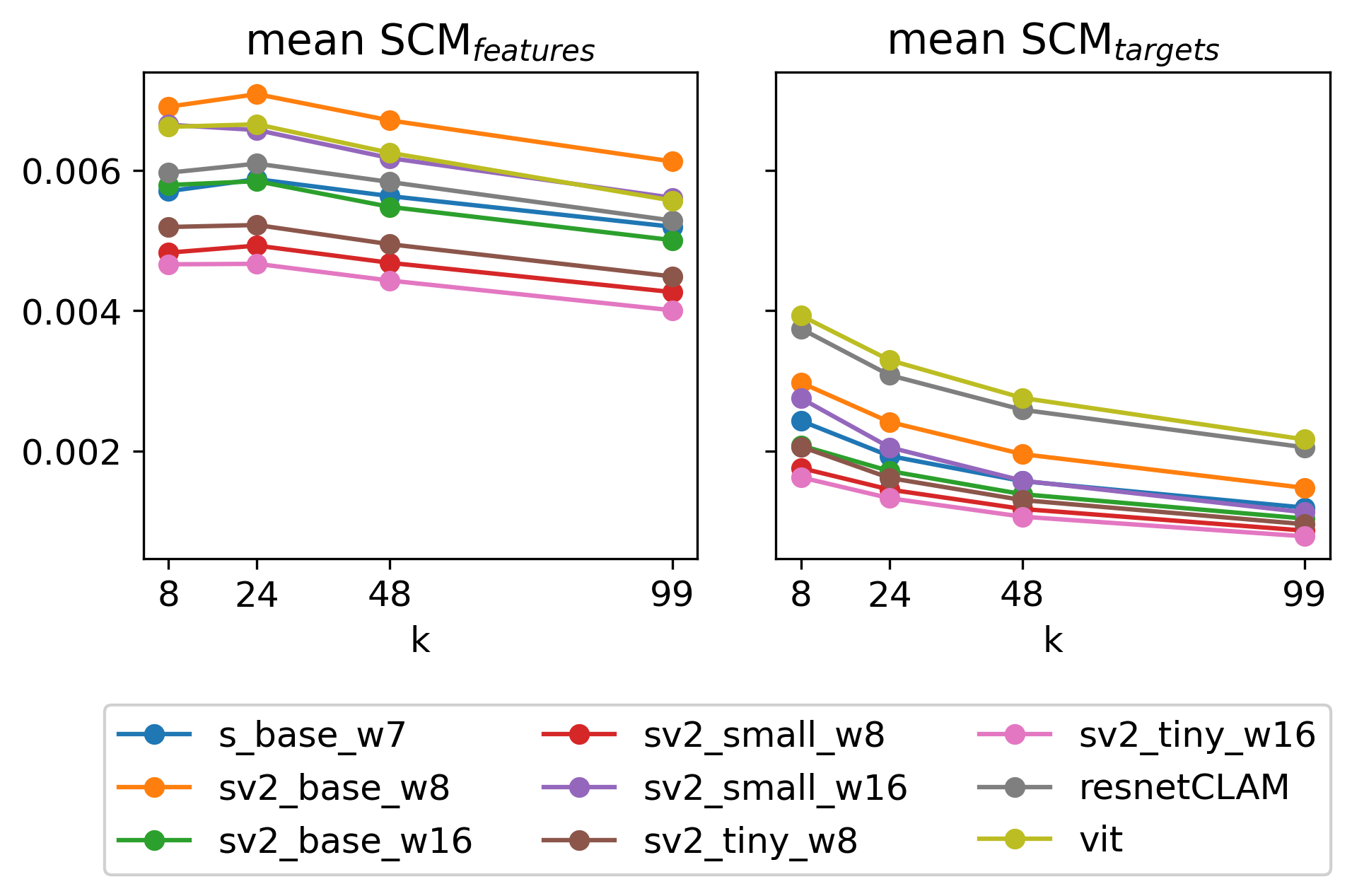}



\caption{The mean value of spatial context measures 
for models with different feature extractors with respect to the size of neighborhood in the spatial weights matrix within spatial regression models over all images with tumor.}
\label{fig:k}
\end{figure}

The analysis of the results reveals that the size of~the~neighborhood $k$ incorporated in the spatial weights matrix within spatial regression models has an impact on the spatial context measures given tumor data. The presented values are the average measures over all tumor images which is the cause of relatively small values (the analogous plots for the maximum values (reaching even more than 0.04) are provided in the Appendix). \\
In~the~case of $SCM_{targets}$, there is a clear decreasing trend (the bigger \emph{k}, the smaller $R^2$ boost), whereas, in the case of $SCM_{features}$, it depends on the feature extractor used (see Figure~\ref{fig:k}). For some DL models, there is an initial very slight increase from $k=8$ to $k=24$ and then the drop, and for others, there is a decreasing trend on the whole spectrum. In general, however, all the models behave in a similar way as the curves are largely parallel. The decreasing trend on the plots aligns with the first law of geography (by Waldo R. Tobler) saying that \textit{`everything is related to everything else, but near things are more related than distant things'}.

\subsubsection{Neighborhood size vs. window size in \textit{Swin v2} models}
The general claim regarding the analyzed models based on~Figure~\ref{fig:k} is that the continuity and spatial correlation of targets (measured using $SCM_{targets}$) occurs more in the~adjacent patches than the distant ones. Therefore, the smallest possible spatial context is observed. The question comes whether it is true that the bigger the window size within the $Swin v2$, the bigger extent to which the spatial relations are captured.
It turns out that both in the case of $SCM_{features}$ and $SCM_{targets}$ the aforementioned relationship is true only for \textit{small} model as the curves corresponding to small backbone with $window\_size=16$ are above the corresponding ones with $window\_size=8$ for all $k$. In~the~case of \textit{base} and \textit{tiny}  models, the opposite is true meaning the smaller \textit{window size} the bigger spatial context measures. Note that the conclusions drawn regarding the two measures are consistent which is an asset.\\

\subsubsection{Neighborhood size vs. measure of tumor spread}

In~the~analysis summarized in Figure~\ref{fig:k}, all tumor data was analyzed together without the distinction of whether the~tumor is widespread or clustered. Let us define the measure of tumor spread as a mean Euclidean distance between patches (of size 224x224) containing tumor cells ($dist_{mean}$). The information is extracted from the binary masks attached to the dataset. 

In Figure~\ref{fig:k} it was shown that the spatial context measures often reach the highest values for the smallest $k$. To further analyze the topic, we investigated whether this rule holds for widespread tumors, as in this case, the global context is potentially more likely to be observed. 

\begin{figure}[h!]
\centering
\includegraphics[,width=0.55\columnwidth]{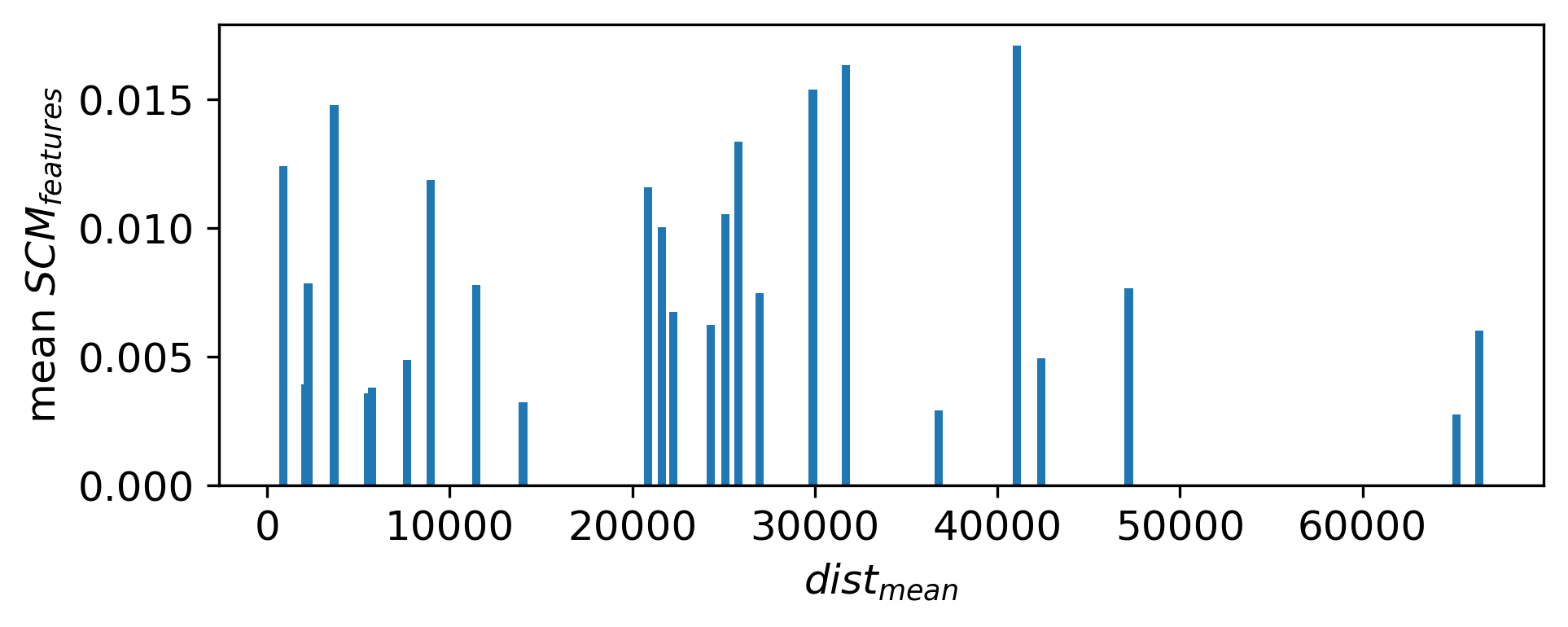}

\caption{Mean of $SCM_{features}$ computed on all tumor images with all model-$k$ combinations where $k=99$ within $W$ is used. The images are characterized by the measure of~tumor spread.}
\label{fig:widespread}
\end{figure}

It turns out that the expected behavior that 
for more widespread tumors, the bigger spatial context measure occurs is not observed (Figure~\ref{fig:widespread}). 
The Pearson correlation coefficient between $dist_{mean}$ and $SCM_{features}$ ranges from 0.070 (for $k=8$) to 0.125 (for $k=99$) which is a monotonical increase with respect to neighborhood size. The~same analysis performed on $SCM_{targets}$ shows that the correlation coefficients are about two times bigger than in the case of $SCM_{features}$ but are less dependent on $k$ - they range from 0.214 ($k=8$) to 0.237 ($k=48$). Therefore, it seems that the~global context even for widespread tumors is not preserved in the models as the correlation rates are low.

%

\subsection{`Most spatial' images}
\begin{figure}[h!]
\centering
\includegraphics[width=0.55\columnwidth]{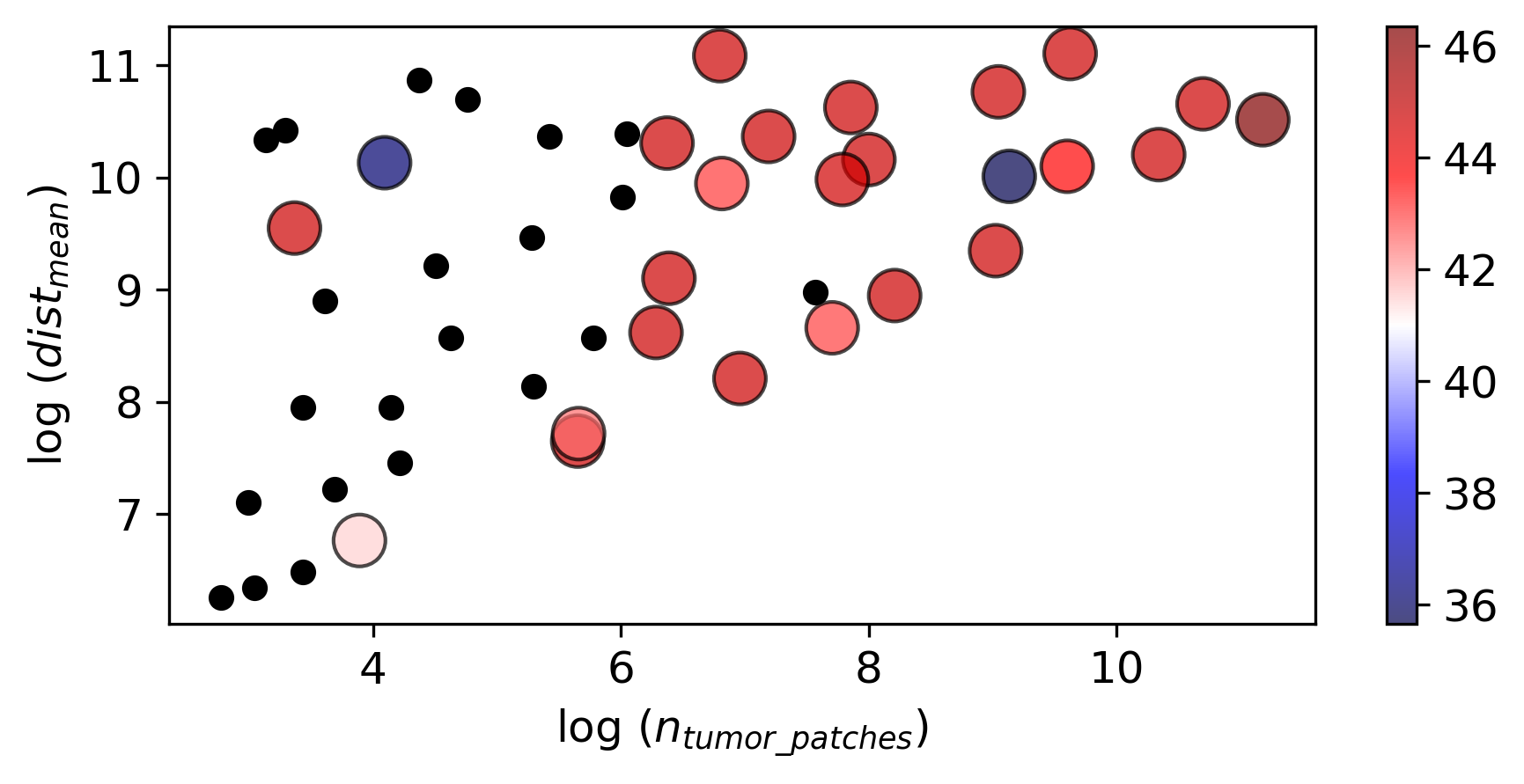}
\caption{The relationship between the logarithms of~$n_{tumor\_patches}$ and $dist_{mean}$. The color denotes $k_{mean}$. The small black dots depict the tumor images that were chosen as `the most spatial' by less than or half of the~model-$k$ combinations. 
}
\label{fig:scatter}
\end{figure}

As already shown in Figure~\ref{fig:density}, $SCM_{features}$ and $SCM_{targets}$ tend to be much bigger for tumor lesions than normal tissues. 
We selected the images with tumor where the spatial context is the most significant. We calculated the threshold as a 95\%-percentile of the $SCM_{lag}$ on the normal lesions. Later, we picked images with tumor with a~larger $SCM_{targets}$ than the threshold. This procedure was performed for each model-$k$ combination separately (results from the folds were aggregated using the mean). For the selected images, we extracted information from their binary masks about (1) the number of tumor patches ($n_{tumor\_patches}$) and (2) tumor spread ($dist_{mean}$ - the same metric as in Figure~\ref{fig:widespread}). 
We applied an additional filter on the images as we picked only the ones that were chosen as `the most spatial' by more than half of all model-$k$ combinations. 
Moreover, we computed the mean value of \emph{k} within the model-$k$ combination that led to the decision that a particular image is the `most spatial' ($k_{mean}$). The values are summarized in Figure~\ref{fig:scatter}.

It is seen that most of the images pointed as `the most spatial', are the ones where there are many patches with tumor cells. The point $log(n_{tumor\_patches})=6$ could be perceived as a cutoff which gives about 400 tumor patches. Note that the fact whether a tumor is widespread or clustered ($dist_{mean}$) does not play a significant role. 



\subsection{Spatial context measures 
vary depending on the~feature extractor used}
\label{sec:R2_boost_tumor}

In order to analyze the differences in DL models with different backbones from the spatial context point of view, we prepared a ranking based on spatial context measures given tumor data (Table~\ref{tab:ranking}). It turns out that the places from 5. to 9. are the same for the mean values of $SCM_{features}$ and $SCM_{targets}$. At the first four places are models of different architectures - \textit{resnetCLAM}, \textit{ViT} and \textit{Swin v2}. The order of the models varies depending on the analyzed spatial context measure. It seems that the architecture type - convolutional vs. transformer-based neither the attention coverage (global - in \textit{ViT} vs. local - in \textit{Swin, Swin v2}) does not directly impact the~measures of spatial context. Within the four first places are mostly models that generate embeddings of the bigger size (1024) - \mbox{\textit{resnetCLAM}}, \textit{ViT} and \textit{Swin v2 base}. The exception is \textit{Swin v2 small}. Note that after dropping the correlated features, the number of predictors in regression models has significantly decreased (the most, more than 2 times, in the case of features generated within \mbox{\textit{resnetCLAM}}).\\
As already seen in Figure~\ref{fig:k}, for \textit{base} and \textit{tiny} models there is a rule that the smaller the window size, the bigger the spatial context measure given the same model architecture (Table~\ref{tab:ranking}). The statistical significance of the differences was analyzed using a pairwise Wilcoxon test with $\alpha =0.05$.

Note that the differences between the spatial context measures within DL models with different backbones 
revealed in the comparative analysis can be directly attributed to the design of the feature extractors and the values of the resulting features because there was no significant difference in the overall classification performance of the DL models (see Figure~\ref{fig:model_perf}).

\begin{table}[h!]
\caption{Ranking of DL models based on the mean spatial context measures. In the brackets, the mean number of predictors (over all test images) used in regression models is given (after dropping the correlated ones). The values are provided only in one column for brevity.} 
\centering
\label{tab:ranking}
\begin{tabular}{cccl}
\toprule
\textbf{ranking} & \textbf{mean $SCM_{features}$}  & \textbf{mean $SCM_{targets}$} \\ 
\midrule
1. & \textit{sv2\_base\_w8} ($736$) & \textit{vit} \\ 
2. & \textit{vit} ($703$) & \textit{resnetCLAM} \\ 
3. & \textit{sv2\_small\_w16} ($572$) & \textit{sv2\_base\_w8} \\ 
4. & \textit{resnetCLAM} ($410$) & \textit{sv2\_small\_w16} \\ 
5. &\textit{s\_base\_w7} ($755$)  & \textit{s\_base\_w7} \\ 
6. &\textit{sv2\_base\_w16} ($747$)  & \textit{sv2\_base\_w16}\\ 
7. & \textit{sv2\_tiny\_w8} ($585$) & \textit{sv2\_tiny\_w8}  \\ 
8. & \textit{sv2\_small\_w8} ($577$) & \textit{sv2\_small\_w8} \\ 
9. & \textit{sv2\_tiny\_w16} ($615$) & \textit{sv2\_tiny\_w16} \\ 
\bottomrule
\end{tabular}
\end{table}




\subsection{Difference in spatial context measures within training and test sets}

Sometimes DL models generalize poorly on the test set which is observed within the performance metrics. The question is whether it is also the case in the ability to capture the spatial context. 
We performed the independent samples t-test 
to verify whether there is a significant difference in mean spatial context measures on tumour lesions within training and test sets. It turned out that in more than 86\% of all model-$k$ combinations, the mean of $SCM_{features}$ and $SCM_{targets}$ were significantly greater on the training set than the test set. Therefore, it seems that DL models do not generalize well in capturing spatial context despite the use of the early stopping mechanism during model training.

\subsection{Local inspection of the features-attention scores relationship}
\begin{figure}[htb!]
\centering
\includegraphics[width=1\textwidth]{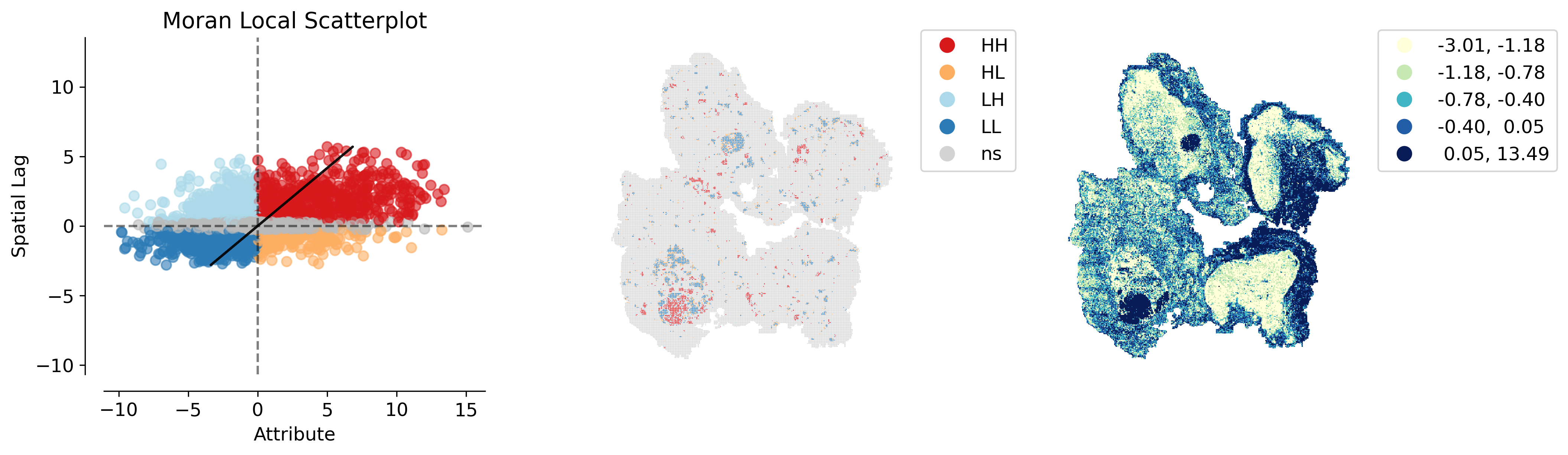}
\caption{On the left, the Moran Local Scatterplot where the residuals from OLS and the spatially lagged counterparts are shown (\textit{sv2\_base\_w8}, \emph{k=24}). The slope of the linear fit corresponds to the Global Moran's I statistic. 
In the center, the Local Spatial Autocorrelation cluster map is provided on OLS residuals (\emph{ns} - not statistically significant clusters, \emph{HH} - means that the high value of the residual is surrounded by high values, \emph{HL} - the high value is surrounded by low values which show negative spatial correlation. Note that in Moran's I statistics, we operate on deviations from the mean therefore the \emph{high} and \emph{low} refer to the values above and below the~average. An analogous LISA analysis could be performed on raw attention scores. On the right, the Choropleth map is provided with the raw attention scores grouped into 5 quantiles.}
\label{fig:one_spatial}
\end{figure}

Spatial statistics allow not only to analyze data from the~global point of view but also from the local one. 
One of the methods is to use a local indicator of spatial autocorrelation (LISA) \cite{lisa} which is formally similar to the Global Moran's I statistic but the summation over all regions is removed. We provide a visualization of one of `the most spatial' images (\textit{`test\_001.tif'}) from the local perspective using LISA (Figure~\ref{fig:one_spatial}). 

\section{Conclusions}
We propose a \textit{`Deep spatial context'} ($DSCon$) method for the investigation of the Deep Learning models from the perspective of spatial context with is a human understandable concept.
In the method, the inspection of the spatial aspect of the relationship between the features extracted from the patches and the assigned raw attention scores within a DL model for tissue classification is performed. 

The \textit{DSCon} method serves as a good framework for the analysis of various research questions related to spatial context in different areas. In our investigation on the illustrative digital pathology use case, we observe that the spatial relationships are maintained by the models mostly in the case of tumor lesions, which is in line with histopathologists' expertise. Furthermore, we noticed that spatial relationships can be retrieved mainly from the local context, $k=8$. The spatial signal is larger within the patch features than within the neighboring attention scores (targets). 
We point out that spatial statistics can serve not only for the analysis of the behavior of attention-based models on the collections of images, but also for the local inspection of single images which can provide more insights into the structure of the lesion.

Although inspired during the collaboration with a histopathologist, the \textit{DSCon} method itself is general and can be applied to various types of datasets with images of huge resolution processed using the attention-based models to study the concept of spatial context.

\bibliographystyle{unsrt} 
\bibliography{main}

\newpage
\appendix
\section{Appendix}
\subsection{Training details}
After an extensive search over a grid of hyperparameters, the following parameters were chosen for the final experiments on models with transformer-based backbones:
\textit{learning\_rate=0.005}, \textit{weight\_decay\_rate=0}, \textit{dropout\_rate=0.25},
\textit{bag\_weight=0.7}. In the first layer of the classifier, there were 512 neurons, whereas 128 in the second. Following the setup in~\cite{vpt}, the SGD optimizer was used. 

\subsection{Global Moran's I test}
 Let us suppose that there is a grid of regions $i=1,..,N$ with the corresponding value $x_i$. For each region, $i$, the cross-product of the deviation from the global mean ($z_i=x_i-x$) and the lagged value of the counterpart is computed.  
 The lagged value can be thought of as an average of the values within the neighbors defined in the spatial weights matrix. 
The formula for the Global Moran's I statistic is as follows:
\begin{equation}
\label{eq:moran}
    I=\frac{\sum_i\sum_j w_{ij}z_i*z_j/S_0}{\sum_i z_i^2/N}
\end{equation}
 where $S_0$ is a sum of all elements $w_{ij}$ within the spatial weights matrix\\

The null hypothesis in the Global Moran's I test is that there is spatial randomness as opposed to spatial autocorrelation. 
To investigate the random factor, Moran's I statistic is calculated on different permutations of regions and the corresponding values within the grid. As a result, a set of Moran's I statistics forms a distribution that is used to compute a `pseudo' $p$-value that is then compared against the~significance level.

\subsection{Impact of size of neighborhood (\emph{k}) 
on the maximum spatial context measures}
\begin{figure}[h!]
\centering
\includegraphics[,width=0.5\columnwidth]{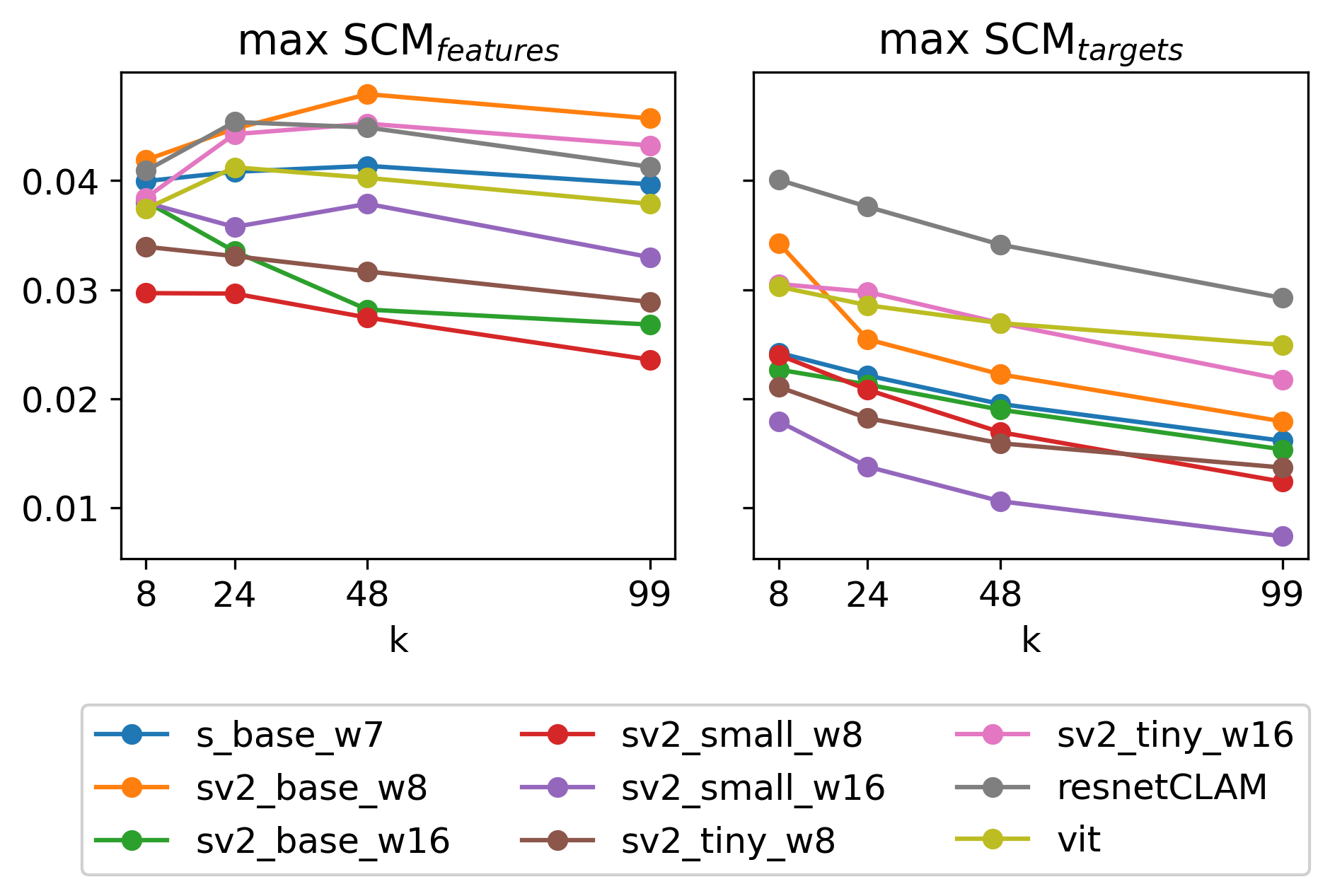}
\caption{The maximum value of spatial context measures for different feature extractors in respect to the size of neighborhood in the spatial weights matrix within spatial regression models over all images with tumor.}
\end{figure}
In the case of maximum measures, the conclusions for $SCM_{targets}$ are the same as for the mean values. However, it is differently in the case of $SCM_{features}$. Here, there is no clear relationship between the curves corresponding to models with different window sizes. This can be attributed to the fact that maximum values are more bias by nature.

\end{document}